\documentclass{article}

\usepackage{PRIMEarxiv}

\usepackage[utf8]{inputenc} 
\usepackage[T1]{fontenc}    
\usepackage{hyperref}       
\usepackage{url}            
\usepackage{booktabs}       
\usepackage{amsfonts}       
\usepackage{nicefrac}       
\usepackage{microtype}      
\usepackage{lipsum}
\usepackage{fancyhdr}       
\usepackage{graphicx}       
\graphicspath{{media/}}     
\usepackage{todonotes}
\usepackage{amsmath}

\usepackage{multirow}
\usepackage{array}
\usepackage{algorithm}
\usepackage{algpseudocode}
\usepackage{pifont}
\usepackage{caption} 

\newcolumntype{C}[1]{>{\centering\let\newline\\\arraybackslash\hspace{0pt}}m{#1}}

\title{A deep learning attention model to solve the Vehicle Routing Problem and the Pick-up and Delivery Problem with Time Windows
\thanks{\textit{\underline{Citation}}: 
\textbf{Baptiste Rabecq, Rémy Chevrier}} 
}

\author{
    Baptiste Rabecq \\
    CentraleSupélec, SNCF \\
    Paris, France\\
    \texttt{baptiste.rabecq@student-cs.fr} \\
    \And
    Rémy Chevrier\\
    SNCF \\
    Paris, France\\
    \texttt{remy.chevrier@sncf.fr}
}

\begin{document}
\maketitle

\numberwithin{equation}{section}

\newcommand{\CVRPTW}{\text{C-VRP-TW}}
\newcommand{\CVRP}{\text{C-VRP}}
\newcommand{\VRPTW}{\text{VRP-TW}}
\newcommand{\CPDPTW}{\text{C-PDP-TW}}
\newcommand{\PDPTW}{\text{PDP-TW}}

\begin{abstract}
    SNCF, the French public train company, is experimenting with developing new transportation services by tackling vehicle routing problems.
    While many deep learning models have been used to tackle vehicle routing problems efficiently, it is challenging to consider time-related constraints.
    In this paper, we solve the Capacitated Vehicle Routing Problem with Time Windows (\( \CVRPTW \)) and the
    Capacitated Pick-up and Delivery Problem with Time Windows (\(  \CPDPTW \)) with a constructive iterative
    Deep Learning algorithm. We use an Attention Encoder-Decoder structure and design a novel insertion heuristic
    for the feasibility check of the \(  \CPDPTW \). Our model yields results that are better than the best-known
    learning solutions on the \( \CVRPTW \). We show the feasibility of deep learning techniques for solving the
    \(  \CPDPTW \)  but witness the limitations of our iterative approach in terms of computational complexity and efficiency.
\end{abstract}

\keywords{Pick-up and Delivery \and Deep Reinforcement Learning \and Vehicle Routing Problem \and Time windows \and Attention}

\section*{Introduction}
SNCF is looking for new ways of building public transportation services with limited exploitation costs and a
better adaptation to demand. Therefore, SNCF is interested in swiftly constructing vehicle tours for dial-a-ride
applications and is now focusing its research on on-demand transportation systems.

The Vehicle Routing Problem is a well-known combinatorial optimization problem, generalizing the Travelling
Salesman Problem.
Given a graph composed of a depot and customer nodes and a fleet of vehicles, we visit all the customers once while
minimizing the total distance made by the fleet.
The problem can be further complexified by adding time windows to the depot and customer nodes to limit the time span of the solutions.
Moreover, we can add precedence constraints to the model to force it to visit some nodes before others.
This problem is called the Pick-up and Delivery Problem with Time Windows (\(  \CPDPTW \)).
In this paper, we design a constructivist end-to-end deep reinforcement learning model to solve the \(  \CPDPTW \).

In our resolution method, a deep learning agent iteratively chooses nodes to visit to update its state.
After every decision, the agent enters a new state, with a new customer added to his route.
The environment provides him with the resulting feasible nodes so that the agent can choose his next customer, and so on.

While other constructivist models have successfully been used to solve Vehicle Routing Problems, the combinatorial complexity added by the time windows 
and the precedence constraint is very challenging.
Considering them in a constructivist learning model has not been shown to be competitive in terms of pure cost against classic heuristics.
However, the method is well-adjusted to give very fast approximate solutions.
Therefore, in this work, we extend the scope of resolution of the Attention Model~\cite{kool2018attention} to more constrained problems.
While keeping a generic learning agent, we modify its computing environment so that precedence constraints and time windows can be considered.

Our contribution, therefore, lies in the environment and the feasibility check of the deep learning model.
We pair the decision model with a heuristic environment, allowing for fast computations even in the case of the Pick-up and Delivery.
More specifically, we compute approximate feasibility with an insertion heuristic that reduces the complexity of the feasibility check from exponential to quadratic.

\section{Related Work}

In the last few years, a lot of work using Deep Reinforcement Learning to solve VRP appeared. 
The first approach, by Nazari et al.~\cite{NEURIPS2018_9fb4651c}, used Pointer Networks~\cite{vinyals2015pointer}. 
However, as Pointer Networks are not invariant to the order of encoding of the node, they were replaced in the work of Kool et al.\ 
by the Attention Model~\cite{kool2018attention},
 which uses a multi-head attention layer in the encoder and a greedy rollout baseline in the REINFORCE\cite{REINFORCE} algorithm. 
 This work was further improved by Peng et al.~\cite{Pengdynamic}, who proposed a dynamic encoder-decoder model for the \( \CVRP \), by re-encoding the nodes 
 with the Attention Model, every time a vehicle comes back to the depot. 
 Moreover, Kwon et al.~\cite{POMO} improved the results of the Attention Model by replacing the greedy rollout baseline with their POMO baseline, 
 which consists in solving multiple times the same instance and using the average of the solutions of the instance as a baseline. 
 This was proven to reduce the variance of the baseline and yielded better results. The POMO baseline reduces computation times, 
 as it removes the pre-computing step needed with the greedy rollout baseline.

However, taking time window constraints into account is very challenging. In 2020 Falkner et al.~\cite{falkner2020learning} proposed JAMPR, 
based on the Attention Model to build several routes jointly and enhance context. 
However, the high computational demand of the model makes it hard to use. Concurrently, Wang et al.~\cite{Hierarchical} proposed a hierarchical 
reinforcement learning model based on pointer networks. 
Their idea is to learn to get feasible solutions on the lower level and then use this solution as input to a second decoder, aiming to get 
solutions minimizing the total distance. While this approach is interesting, it shows good results only with a few customers (around 20).

Moreover, Bono~\cite{multi-agent} develops MARDAM, a multi-agent model to tackle Stochastic \( \CVRPTW \). 
More recently, using this same approach, Sultana et al.\ used a multi-agent model to tackle the stochastic \( \CVRPTW \)~\cite{Fastapproximate}. 
They got also yielded great results on the static \( \CVRPTW \) as well on the Solomon dataset~\cite{Solomon}.

The Pick-up and Delivery Problem has also been studied by Li et al.~\cite{Heterogeneous}, using Heterogeneous Attention to encode the non-symmetric 
relations between pick-up nodes and delivery nodes. Lowens et al.\cite{TSPprecedence} have further adapted this work using the sparse attention mechanism
 to solve the TSP with precedence constraints. It reduced the complexity of the calculations, hence, improving their computation time.

\paragraph{Contributions}
In this work, we extend the solving capabilities of the  Attention Encoder-Decoder model for the resolution of the Capacitated Pick-up and Delivery Problem with 
Time Windows (\(  \CPDPTW \)) and the Capacitated Vehicle Routing Problem with Time Windows (\( \CVRPTW \)). 
Specifically, we change the environment of the agent according to the problem we solve.
We show that the feasibility check of the \(  \CPDPTW \)  is a lot more computationally demanding than the \( \CVRPTW \).
We develop a new environment based on an insertion heuristic for the Pick-up and Delivery with Time Windows.
In section 2, we provide definitions for the problems we solve. Then, in section 3, we describe the Attention model we used in our resolution. 
In section 4, we describe the insertion heuristic coupled with the decision model. Finally, we provide our results on the \( \CVRPTW \) and the \(  \CPDPTW \).

\section{Problem definition}
\subsection{\texorpdfstring{\( \CVRP \), \( \CVRPTW \), \(  \CPDPTW \)}{C-VRP, C-VRP-TW, C-PDP-TW} }

In this paper, we solve the Capacitated Pick-up and Delivery with Time Windows. We show how, starting from a \( \CVRP \), we can add constraints to get the \(  \CPDPTW \).

\paragraph{Vehicle Routing Problem}

We use the definition of the VRP of Desaulniers et al.~\cite{desaulvrp}. Let $G = (V, E)$ be a Euclidean graph, with $|V| = N+2$, $N$ customer nodes, 
and a depot node, represented by indices $0$ and $N+1$. Each node $i \in \{0, 1, \dots, N+1\}$ has three components: its coordinates $x_i$, $y_i$, and its demand $q_i$. 
We consider that $q_0 = q_{n+1} = 0$.

We consider a fleet of vehicles of size K, each with a capacity $c_k$. We seek to allocate these vehicles to the different nodes in such a way as to minimize 
the total distance traveled, under the constraint of not exceeding the vehicles' capacity. We will iteratively construct routes $\tau_k \in {N}^{N+2}, k\in \kappa$, such that:
\begin{align}
    \forall j \ne k,\quad   & \tau_j \bigcap \tau_k = \{0, N+1\}                                                        \\
                            & \bigcup\limits_{i \in \kappa} \tau_k = V                                                  \\
    \forall j \leq K, \quad & \tau_j[0] = 0                                                                             \\
    \forall j \leq K, \quad & \forall i,\; 1 \leq i \leq N+1,\quad \tau_j[i] \ne 0                                      \\
    \forall j \leq K, \quad & \forall i \leq N+1, \quad \tau_j[i] = N+1 \implies \forall l \geq i,\quad \tau_j[l] = N+1 \\
    \forall j \leq K, \quad & \quad \sum_{i \in \tau_j} q_i < c_j
\end{align}
Each edge $(i, j)$ of the route $\tau_k$ connecting point $i$ to point $j$ is called $e_{k,i,j}$. We call $E_k$ the set of indices of the edges of the route $\tau_k$.
We call $c_{k, i}$ the remaining capacity of vehicle $k$ at point $i$. We need to check that $\forall i \in \tau_k, c_{k,i} > 0$,
We seek to minimize the objective
\begin{equation}
    \sum_{k=1}^{K}\sum\limits_{(i, j) \in E_k} c_{i,j}
\end{equation}

\paragraph{Adding time windows}
In the case of a \( \CVRPTW \), each node has, in addition, a time window $[a_i, b_i]$ in which the vehicle must arrive and a service time $\tau_i$. 
The total time of the instance is given by $[E, L] = [a_0, b_0] = [a_{n+1}, b_{n+1}]$. We set $\tau_0 = \tau_{n+1} = 0$, and we ensure that the problem is feasible by setting:
\begin{equation}
    b_0 \geq \max_{i \in V \backslash \{0\}}(a_0 + \max(t_{0,i}, a_i) + \tau_i+ t_{i,n+1})
\end{equation}
\begin{equation}
    a_0 \leq \min_{i \in V \backslash \{0\}} (b_i - t_{0,i})
\end{equation}

\paragraph{Pick-up and Delivery}
We use the definition given by Batarra et al.~\cite{batarrapdptw}. 
Let $G = (V, E)$ be a Euclidean graph, with $|V| = 2N+2$, $2N$ customer nodes, and a depot node, represented by indices $0$ and $2N+1$. 
The nodes $n$ such that $0\leq n \leq N$ are pick-up nodes, whereas the nodes such that $N+1\leq n \leq 2N+1$ are delivery nodes.
Each node $i \in \{0, 1, \dots, 2N+1\}$ has six components: its coordinates $x_i$, $y_i$, its demand $q_i$, its time window $[a_i, b_i]$ 
and its service time $s_i$. We consider that $q_0 = q_{2n+1} = s_0 = s_{2n+1} = 0$.

The main constraints are very similar to the \( \CVRPTW \), but we add the precedence constraint. Therefore, we aim to build tours $\tau_k \in {N}^{2N+2}, k\in \kappa$, such that:
\begin{align}
    \forall j \ne k,\quad   & \tau_j \bigcap \tau_k = \{0, 2N+1\}                                                         \\
                            & \bigcup\limits_{i \in \kappa} \tau_k = V                                                    \\
    \forall j \leq K, \quad & \tau_j[0] = 0                                                                               \\
    \forall j \leq K, \quad & \forall i,\; 1 \leq i \leq 2N+1,\quad \tau_j[i] \ne 0                                       \\
    \forall j \leq K,\quad  & \forall i \leq 2N+1,\quad \tau_j[i] = 2N+1 \implies \forall l \geq i,\quad \tau_j[l] = 2N+1 \\
    \forall i\leq N,\quad   & \forall l\leq 2N+1,\quad \tau_j[l] = i \implies \exists m > l, \quad \tau_j[m] = N+i
\end{align}

Although the optimal solutions only consider a length penalization, heuristics usually minimize the number of vehicles first and then the total distance of the solution.
In the case of the Pick-up and Delivery, that is what we are trying to solve.

\subsection{Learn to optimize}

The resolution of a vehicle routing problem can be modeled by a sequential decision-making problem, where a customer node is chosen by an agent at every time step. 
To make the decision, we use an Encoder-Decoder structure that sequentially computes the probability of choice of the non-visited nodes.

To allow the resolution with a Reinforcement Learning model, we need to define the states visited and actions taken by our agent. 
Naturally, the actions are the selection of the customer nodes of the graph, $i$.

The states are the partial solutions already constructed. We name $\tau$ a full trajectory, i.e., a complete solution to our optimization problem. 
Similarly, $\tau_t$ is a partially constructed trajectory with $t$ nodes. $f$ is the cost function used, here equal to the total distance of the solution.

Given a partially constructed route $\tau_t$, we compute the following state $\tau_{t+1} = (\tau_{t}; i)$ where $(\cdot\,\cdot)$ is the concatenation operator 
by choosing $i^{(t+1)}$ using the learned function $\pi(\tau_{t+1}, \tau_t, \theta) = \mathbb{P}(\tau_{t+1} = (\tau_{t}; i) | \tau_t;\theta)$, 
where $\theta$ are the parameters of the model.
Using the chain rule, we can infer the probability of choice of a route of length $I$ with:
\begin{equation}
    \pi(\tau|\theta) = \prod_{i=1}^I \mathbb{P}(\tau_i|\theta; \tau_{i-1})
\end{equation}
Therefore, the sequential construction of the solution allows for a Markov Decision Processus modelization and for solving with a Deep Reinforcement Learning framework.

The objective of this selection is to minimize the expected cost of the solution. In other words, we want to find
\begin{equation}
    \pi = \mathrm{argmin}\left(\mathbb{E}_\theta(\pi(\tau|\theta) * f(\tau))\right)
\end{equation}

\section{Deep Learning agent}

Our solving model is composed of a decision model using self-attention layers and a computing environment. 
Given an instance, the model iteratively chooses nodes to visit and updates its state.

\subsection{Attention Model}

The AM model, by Kool et al.~\cite{kool2018attention} uses the attention mechanism to get rid of dependence on the encoding order of the nodes, 
as was the case with RNNs and Pointer Networks.

\subsubsection{Encoder}
First, we project the attributes $x_i$ of each node onto a space of dimension $d_{emb}$. 
The dimension $d_{in}$ of $x$ is 3 for \(  \CVRP \) (position, demand) and 5 for \( \CVRPTW \) (position, demand, time window).
\begin{equation}
    \forall i \in \{0, \dots, N\}, z_i^0=W_0x_i + b_0.
\end{equation}
where $W_0 \in \mathbb{R}^{d_{in} \times d_{emb}}$ and $b_0 \in \mathbb{R}^{d_{emb}}$.

We then apply successively several blocks of self-attention $SA$ to the embeddings of nodes.
\begin{equation}
    z_i^{l+1} = SA(z_i^{l}, Z^{l}), \forall i, l \in V\times \{1, \dots, L\}
\end{equation} where $Z^l = \{z_0^{l}, \dots,z_N^{l}\}$ is the concatenation of the embeddings at the output of layer $l$, and $L$ is the number of layers of the encoder, 
usually chosen between 3 and 6.

Each block is composed of a multi-head attention layer, a dense feed-forward layer with Relu activation (FF), each followed by a residual connection (res) and a 
normalization: batch norm or instance norm (Norm). We thus have:
\begin{equation}
    SA(z_i^{l}, Z^{l}) = Norm(FF^{res}(Norm(MHA^{res}(z_i^{l}, Z^{l}))))
\end{equation}
We define, as usual, the networks as:
\begin{equation}
    FF(x; W, b) = \max(0, Wx + b)
\end{equation}
\begin{equation}
    FF^{res}(x) = x + FF(x)
\end{equation}
\begin{equation}
    MHA^{res}(x) = x + MHA(x)
\end{equation}

The multi-head attention is a linear combination of a single-head attention layer $SHA$ over $num_{heads} = 8$ slices of nodes. (We omit the $l$ and $i$ for readability).

\begin{equation}
    MHA(z, Z; W) = \sum_{k=1}^{num_{heads}} W_k^{heads}SHA(z_{slice(k)}, Z_{.,slice(k)}; W)
\end{equation}
where W represents the parameters of the SHA layer. $W = (W^{heads}, W^{key}, W^{query}, W^{value})$

A SHA layer is defined by:
\begin{equation}
    SHA(z, Z; W) = \sum_{i=1}^N \text{attn}{(z, Z;W^{query}, W^{key})}_i,W^{value}Z_i
\end{equation}
\begin{equation}
    attn(z, Z;W^{query}, W^{key})=softmax(\frac{1}{\sqrt{d_{key}}}z^T{(W_{query})}^T W_{key} Z_{i |i\in \{1, \dots, N\}}), z \in Z
\end{equation}
where $d_{key}$ is the height of the key: $d_{key}=\frac{d_{emb}}{num_{heads}}$

At the output of the encoder, we thus get:
\begin{equation}
    \forall i \in \{0, \dots, N\}, \quad\omega_i = \underbrace{SA\circ\dots\circ SA(z_i, Z)}_{\text{L times}}
\end{equation}

\subsubsection{Context}
The strength of the Attention Model lies in its adaptability to the current situation of the graph. 
Indeed, at each iteration, the decoder is provided, in addition to the encoded nodes, with a dynamically encoded context that allows 
it to adapt to the partial solution it has already constructed. 
For the \( \CVRP \), the context is composed of the last visited node, the current capacity of the vehicle, 
and the average of the embeddings of the nodes, 
called graph embedding $\omega_{graph} = \frac{1}{N+1}\sum\limits_{i=0}^{N} \omega_i$.
At each iteration, we provide the decoder with the context:
\begin{equation}
    \omega_{context} = \omega_{graph} + W_{proj}(\omega_{last\_node}; c)
\end{equation}
where $(\cdot\,;\cdot)$ is the concatenation operator and $W_{proj} \in \mathbb{R}^{d_{emb}+1\times d_{emb}}$ and $c$ is the residual capacity of the vehicles.

\subsubsection{Decoder}
The principle of the decoder is quite similar to the encoder. We compute a query, a key, and a value from our context and the embedding of our current node. 
Then we decode it with the attention mechanism by masking the visited nodes.

The probability $p_i = \pi(\tau_{t+1}, \tau_t, \theta)$ of choosing node $i$ at time step $t+1$ is:
\begin{equation}
    p_i = \text{attn}(\text{MHA}(\omega_{context}, W), mask(M))
\end{equation}

\subsection{REINFORCE algorithm with POMO baseline}

We use Policy gradient learning, with the algorithm REINFORCE with baseline invented by Williams\cite{REINFORCE}.
Instead of the greedy rollout baseline used in the work of Kool et al.\cite{POMO}, we use the POMO baseline by Kwon et al., 
which introduces less variance while reducing the computational costs of the model.
Following POMO, we also remove the significance test triggered when the model is improved. 
The pseudo-code is available in algorithm~\ref{alg: pomo}.

\begin{algorithm}
    \caption{REINFORCE with POMO baseline}\label{alg: pomo}
    \hspace*{\algorithmicindent} \textbf{Input:} batch size $B$, number of epochs $E$, number of training steps $T$, number of POMO samples $N$
    \begin{algorithmic}[1]
        \State Init $\theta$
        \For {$epoch = 1, \cdots  E$}
        \For {$STEP = 1,\cdots T$}
        \State $\tau_i \gets RandomInstance() \forall i \in\{1, \ldots, B\}$
        \State $(\alpha_i^1, \ldots, \alpha_i^N) \gets SampleStartNodes(si)\quad \forall i \in\{1, \ldots, B\}$
        \State $\pi_i^j \gets SampleRollout(\alpha_i^j, \theta, \tau_i) \quad \forall i \in\{1, \cdots, B\}, \quad \forall j \in\{1, \cdots, N\}$
        \State $b_i = \frac{1}{N} \sum_{j=1}^N L(\pi_i^j) \quad \forall i \in\{1, \cdots, B\}$
        \State $\nabla_\theta J(\theta) \gets \frac{1}{BN} \sum_{i=1}^B \sum_{j=1}^N (L(\pi_i^j) - b_i)\nabla_\theta\log(p_\theta(\pi_i^j))$
        \State $\theta \gets Adam(\theta, \nabla_\theta J)$
        \EndFor
        \EndFor
    \end{algorithmic}
\end{algorithm}

\subsection{Model overview}
\begin{figure}
    \centering
    \includegraphics[width=0.9\columnwidth]{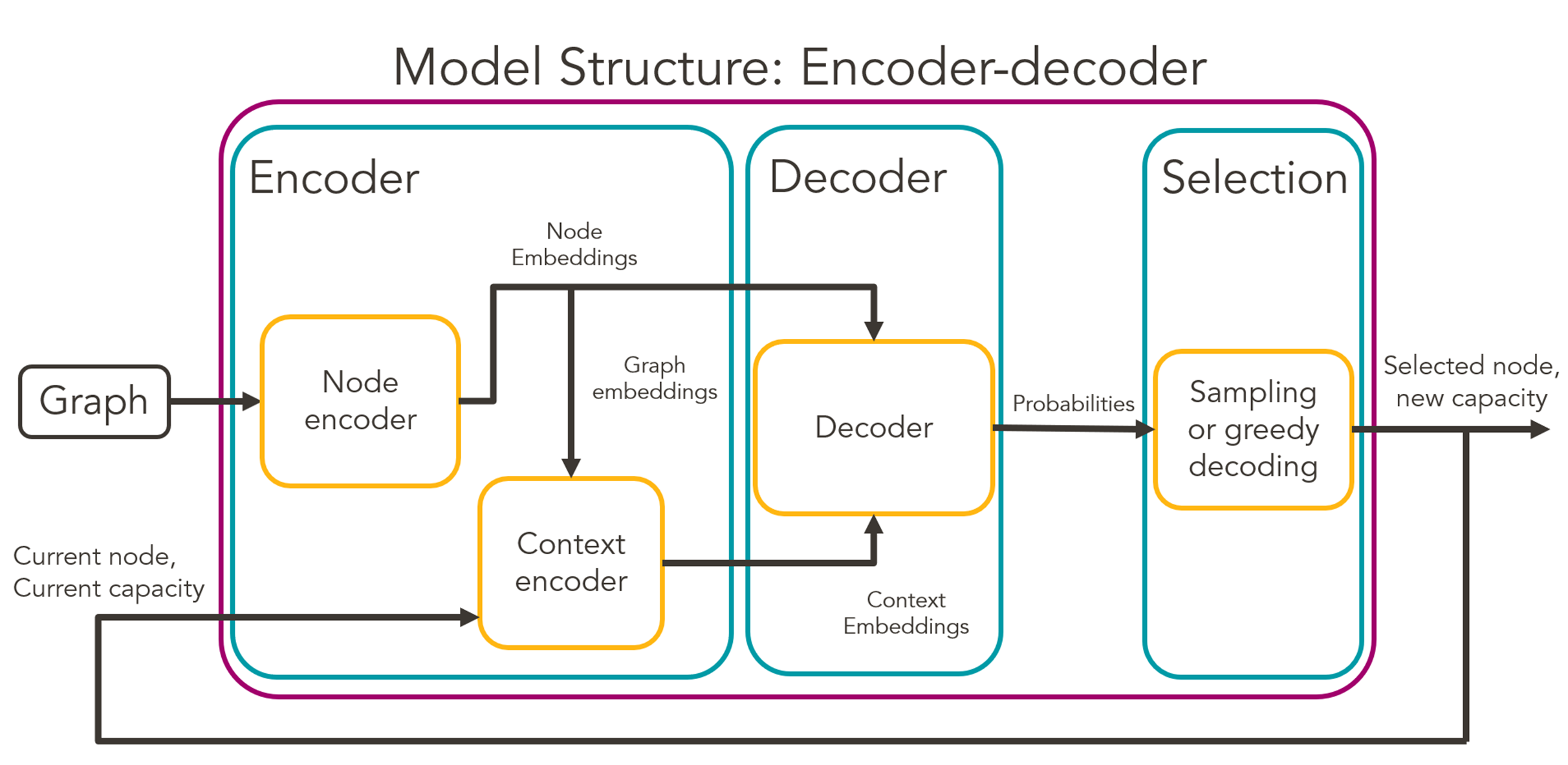}
    \caption{Structure of the deep learning model}
    \label{fig: AM}
\end{figure}

Overall, the model we use is described in figure~\ref{fig: AM}. It is composed of three distinct parts: the encoder, the decoder, and the selection process. 
Once a full solution is constructed, we retro-propagate in the network with the REINFORCE algorithm.

\section{Computing environment}

For the \( \CVRPTW \), the feasibility check is straightforward, as it only involves computing a single

\subsection{Intuition on the issue}

Although adding a precedence constraint looks pretty straightforward, combining it with the time windows constraint severely increases the complexity of the feasibility algorithm. 
The time window constraints imply that a pick-up node cannot be added if we are not sure that we will be able to reach its corresponding delivery node. 
As a consequence, the time window check, which only involved one computation before, is now much more expensive.

\begin{figure}
    \centering
    \includegraphics[width=0.7\columnwidth]{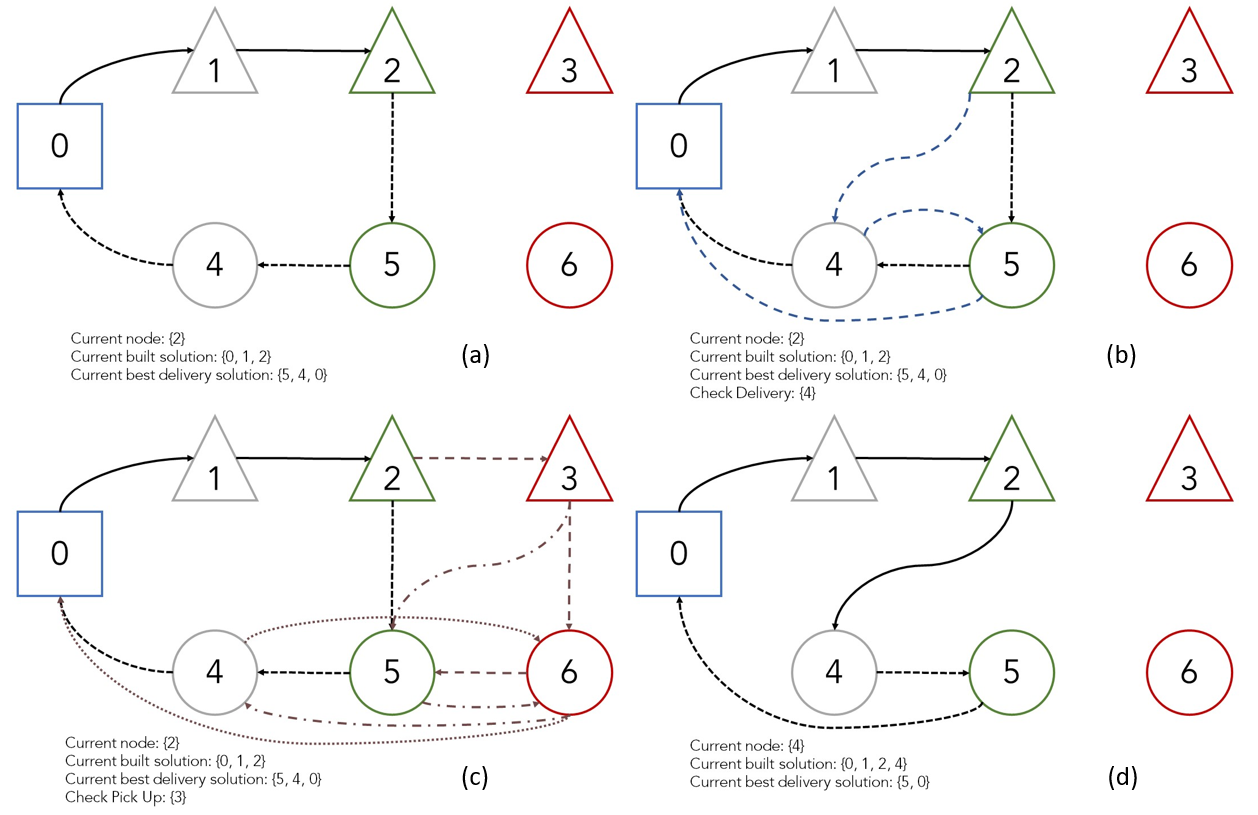}
    \caption{Insertion procedure for the Pick-up and Delivery Problem}
    \label{fig: insertion}
\end{figure}

To illustrate this complexity, we solve "by hand" an instance with three pick-up nodes and three depot nodes in figure~\ref{fig: insertion}. 
To insert a pick-up node, we must check whether it is possible to insert the pick-up and its delivery consecutively. 
To fix ideas, we consider that we have inserted nodes 1 and 2 (figure~\ref{fig: insertion} (a)). 
Therefore, their deliveries, the nodes 4 and 5 have to be inserted right after it. We need to store this information in memory. 
We keep in memory the best solution obtained with the best insertion of the delivery (more information on this later on). 
The environment computed $\{5, 4, 0\}$.

When at point 2, we know that we can insert node 5, as it is the following node in the best-known solution. 
We need to check whether we can insert node 4. 
We will only try to put node 4 in front of the best prospective solution while keeping the rest of the solution as is (figure~\ref{fig: insertion} (b)).
To check on the possible insertion of node 3, we need to insert its delivery, node 6, somewhere in the prospective best-known solution. 
Computing the exact prospective best solution is equivalent to solving the shortest Hamiltonian path between the nodes 5, 6, 4, and 0. 
As this problem is an NP-Hard problem, we cannot afford to compute it at every iteration. 
In order to keep the computing times reasonably low, we use an insertion heuristic, which will only provide approximate feasibility (see algorithm~\ref{fig: insertion}).

In the end, the agent chooses among the nodes that were found feasible. In this particular case, it chose node 4 (figure~\ref{fig: insertion} (d)).

Every time we visit a new pick-up node, we store the prospective solution of deliveries, i.e., the order of the remaining nodes that we have to visit.
 An important distinction has to be made between the constructed solution and the best prospective delivery solution.
The constructed solution $\tau_{1:v}$ is the list of nodes that have already been visited. 
It cannot be permuted whatsoever and is only increased by adding more nodes at the end of it. 
The best prospective delivery solution $\gamma_{1:w}$ is the list of nodes that we know we have to visit but have not been visited yet. It is likely to be changed. 
No node can be removed, but some permutations can be done.

As there are no constraints on the length of travel for the \(  \PDPTW \) (contrary to the Dial-A-Ride-Problem), the best scheduling of the tasks is "at the earliest". 
Therefore, we want to compute the earliest service times $t$ of the nodes we visit or plan to visit.

\subsection{Feasibility check}
In this section, we provide the pseudo-code for the insertion heuristic we developed for the feasibility check of the insertion of a pick-up node and its delivery.
The complexity of this algorithm is proportional to the length of the prospective solution, with a quadratic pre-processing step.

We call $\tau_{1:v} \in \{0, 2N+1\}^t$ the current route that we have visited, $\gamma_{1:w} \in \{N+1, 2N+1\}^t$ the best found route of 
feasible deliveries encountered. $\tau_v$ is the current node while $\gamma_1$ is the first following node.
We call $U \subset \{1, N\}$ the set of non-visited pick-up nodes. $d_{ij}$ is the travel time matrix between the nodes $i$ and $j$.

The pre-processing step consists of computing cumulative waiting times $w_{i,j}, (i, j) \in  \{0, 2N+1\}^2$, slack variables $F_i, i \in  \{0, 2N+1\}$, 
and earliest service time $t_i, i\in \{0, 2N+1\}$ of the nodes $i$ in $\tau_{1:v}$ and $\gamma_{1:w}$. We name $\Delta = \{\tau_v, \gamma_1, \dots, \gamma_w\}$.
\begin{align}
    \forall i, j,  i < j \leq w+1, \quad & w_{\Delta_{i},\Delta_{j}} = \sum_{i < p \leq j}t_{\Delta_{p}} - (t_{\Delta_{p-1}} + d_{\Delta_{p-1}, \Delta_{p}}) \\
    \forall i \leq w+1, \quad            & w_{\Delta_{i},\Delta_{i}} = 0,                                                                                    \\
    \forall i \leq w+1, \quad            & F_{\Delta_i} = \min_{i \leq j \leq w+1} w_{\Delta_i,\Delta_j} + ({b_{\Delta_j} - t_{\Delta_j}})
\end{align}

This pre-processing step is made every time we update the solution by adding a new node to $\tau$ or/and $\gamma$. 
So our Pick-up and Delivery adaptation is split into two distinct parts: first, the update that pre-computes the variables and updates the state of the agent. 
The second step is the feasibility check, which uses these variables to compute the constraints of the environment.

On algorithm~\ref{alg: insertion}, we provide our pseudo-code for the insertion algorithm, adapted from Gschwind et al.\cite{PDPTWfeasibility}. 
Let's stress that the parallel compatibility of the tasks makes the implementation very challenging. 
In particular, it is not conceivable to use for loops in the code, and we need to take into account the fact that the tour lengths can be different. 
Specifically, the computation of the cumulative waiting times is complex. To do so, we use pre-computed lower triangular matrices.

\begin{algorithm}
    \caption{Best insertion algorithm for feasibility checking}
    \label{alg: insertion}
    \hspace*{\algorithmicindent} \textbf{Input:} partially constructed pick-up solution $\tau_{1:v}$, partially constructed delivery solution $\gamma_{1:w}$, set $U$, 
    current node $i$.
    \begin{algorithmic}[1]
        \For {$u$ in $U$}
        \State $t' \gets \max(t_{\tau_t} + d_{iu}, a_{u})$
        \Comment{Compute the simulated service time of node $u$}
        \If{$t' > b_{u}$}
        \State return \verb!False!
        \EndIf
        \State $\delta_{\gamma_1} = t' + d_{u, \gamma_1} - t_{\gamma_1}$
        \Comment{Compute time shift caused by the insertion of the pick-up node $u$}
        \If{$\delta_{\gamma_1} > F_{\gamma_1}$}
        \State return \verb!False!
        \EndIf
        \For {$x = 1,\; \dots, \; v-1$}
        \State $t_{\gamma_x}'' \gets t_{\gamma_x} + {(\delta_1 - w_{\gamma_1, \gamma_x})}^+$
        \State $t_{u+N}'' \gets \max(a_{u+N}, t_v'' + d_{\gamma_x, u + N})$
        \If{$t_{u+N}'' > b_{u+N}$}
        \Comment{Check every possible delivery node new earliest time}
        \State return \verb!False!
        \EndIf
        \State $\delta_{\gamma_{x+1}} \gets (t_{u+N}'' + d_{u+N, \gamma_{x+1}} - t_{\gamma_{x+1}})$
        \Comment{Compute the impact of insertion of the delivery node}
        \If{$\delta_{\gamma_{x+1}} > F_{\gamma_{x+1}}$}
        \State return \verb!False!
        \EndIf
        \State $\text{isFeasible}(u, x)$ = \verb!True!
        \EndFor
        \EndFor
    \end{algorithmic}
\end{algorithm}

In addition to the feasibility check algorithm, we compute the best insertion position by comparing the added cost of insertion of any feasible 
node in the current prospective best solution.

To improve the precision of the algorithm without hurting its computational performance, we want to check the insertion of a node that 
is in the prospective solution but not in the first spot (like in figure~\ref{fig: insertion}(b)).

\section{Training and inference setup}
\subsection{\texorpdfstring{\(  \CVRPTW \) }{C-VRP-TW} with soft and hard time windows}

Several encoder-decoder setups have been tried to tackle the \( \CVRPTW \) efficiently. First, we have tried the multi-agent approach by Falkner et al.~\cite{falkner2020learning}, 
with the joint construction of tours and context enhancement. 
We found that the added computational costs and the added memory usage were not worth it, considering the low improvement compared to the original masked Attention Model. 
Secondly, the dynamic encoding proposed by Peng et al.~\cite{Pengdynamic} was shown to improve the results very slightly while significantly increasing the computational costs.
Therefore, to limit the complexity of the model and allow a pick-up and delivery adaptation, we solve the \(  \PDPTW \) with an almost straightforward 
adaptation of the Attention Model with POMO baseline.

We solved the \( \CVRPTW \) with soft constraints and hard constraints. For the hard constraints, we masked the unfeasible nodes before every decoding step. 
For the soft constraints, we added a strong penalization to the vehicles with late arrival to the nodes.

We test our model on the Solomon dataset with 50 customers. 
We make no distinction between the types of time windows, their frequencies, or the type of instance of the dataset during training because we want our model 
to be as general as possible.
Therefore, for training, we use the instance generator by Falkner et al.~\cite{falkner2020learning} to build our solutions.

\subsection{\texorpdfstring{\(  \CPDPTW \) }{C-PDP-TW}}
As the hard time windows lead to better results on the \( \CVRPTW \), we only used them on the Pick-up and Delivery with Time Windows.
To consider the precedence constraint, we add a feature to every node, -1 if the node is a pick-up node and 1 if it is a delivery. 
Moreover, we encode every pick-up with its delivery to stress the relationship between the nodes and every delivery separately.

We test our model on the Li and Lim dataset~\cite{LiLimPDPTW}, with 100 tasks (50 clients). 
We report the performance of our model using a cost function that tries to minimize the number of vehicles used first and then the total distance made.

\subsection{Model hyperparameters}
We trained our model on instances of 50 customers with 16 distinct starting POMO nodes. We used a batch size of 80 instances. 
With POMO, it represents $80 \times 16 = 1280$ instances per batch.
We trained the model over 200 epochs, each epoch having 20000 instances. 
One epoch is approximately 20 minutes for a \( \CVRPTW \) with 50 customers. We do not use the data augmentation of POMO for inference.

Our initial embedding dimension is 128.
We train our model with 3 attention layers, each of them with a head dimension of 16 and a feed-forward layer of dimension 512. 
According to POMO\cite{POMO}, we use instance generalization instead of the original batch normalization of the Attention Model.
We use a learning rate of $10^{-4}$.

\section{Results}

\subsection{\texorpdfstring{\(  \CVRPTW \) }{C-VRP-TW}}
The results of our model are visible in table~\ref{tab: results}.
We found that the hard constrained problem with the distance as the only objective provides quite good solutions. 
In terms of total distance, our solution yields better results than other learning-based techniques on the Solomon dataset. 
We managed to use one single model producing relatively good solutions to any instance of the Solomon dataset, which are very distinct from one another. 
We found that training on specific types of instances significantly improves the results of the inference on this type of instance, but our model is also 
able to learn from distinct distributions.

However, the model tends to use more vehicles than its competition for the same tasks. 
This is partly due to a misunderstanding of the geometry of the graph. It tends to leave customers out of its tours to finally get to them via a one-customer tour.
We think that using an optimization heuristic, like 2-opt, on a few steps would help solve this issue quite fast. 
However, we do not manage to improve further with the learning model itself.

Furthermore, our model hardly adapts to distinct distributions without loss of performance. 
For example, it has not been able to fully grasp the structure of the clustered graphs of the Solomon dataset when training on different instances.

We did not find any easy way of improving generalization while keeping the model computationally efficient.

\begin{table}
    \centering
    \def\arraystretch{1.5}
    \begin{tabular}{cccccc}
        \hline
        ~                                               & \textit{TW type}       & \textit{Instance type} & \textit{Total distance} & \textit{Gap (\%)} & \textit{Number of vehicles} \\ \hline
        \multirow{3}{4em}{\textit{Our model}}           & \multirow{3}{4em}{1-2} & R                      & 845.3                   & 10.33             & 8.51                        \\
        \cline{3-6}
                                                        &~                       & C                      & 551.65                  & 52.52             & 7.89                        \\
        \cline{3-6}
                                                        &~                       & RC                     & 835.14                  & 14.35             & 7.78                        \\
        \hline
        \multirow{3}{4em}{\textit{DRL baseline}}        & \multirow{3}{4em}{1-2} & R                      & 982.41                  & 28.23             & 8.75                        \\
        \cline{3-6}
                                                        &~                       & C                      & 407.41                  & 12.64             & 5                           \\ \cline{3-6}
                                                        &~                       & RC                     & 951.03                  & 30.22             & 8.13                        \\ \hline
        \multirow{3}{4em}{\textit{Best known solution}} & \multirow{3}{4em}{1-2} & R                      & 766.13                  & 0                 & 7.75                        \\
        \cline{3-6}
        ~                                               &~                       & C                      & 361.68                  & 0                 & 5                           \\
        \cline{3-6}
        ~                                               &~                       & RC                     & 730.31                  & 0                 & 6.5                         \\
        \hline 
    \end{tabular}
    \caption{Results for the \( \CVRPTW \) with 50 customers on the Solomon dataset}
    \label{tab: results}
\end{table}

\subsection{\texorpdfstring{\(  \CPDPTW \) }{C-PDP-TW}}

\begin{table}
    \centering
    \def\arraystretch{1.5}
    \begin{tabular}{|c|C{25mm}C{25mm}|C{25mm}C{25mm}|}
        \hline
        \textit{Instance Name}  & \textit{Number of vehicle (ours)} & \textit{Total distance (ours)}& \textit{Number of vehicles (best known)}  & \textit{Total distance (best known)} \\ \hline
        lr101                    & 25                               & 3005                          & 19                                        & 1650                       \\ \hline 
        lr102                    & 26                                & 3039                          & 17                                        & 1487                       \\ \hline 
        lr103                    & 19                                & 2938                          & 13                                        & 1292                       \\ \hline 
        lr104                    & 15                                & 2994                          & 9                                        & 1013                       \\ \hline 
        lr105                    & 20                                & 3159                          & 14                                        & 1377                       \\ \hline 
        lr106                    & 23                                & 3015                          & 12                                        & 1252                       \\ \hline 
        lr107                    & 19                                & 2854                          & 10                                        & 1111                       \\ \hline 
        lr108                    & 17                                & 2816                          & 9                                        & 969                       \\ \hline 
        lr109                    & 17                                & 2930                          & 11                                        & 1208                       \\ \hline 
        lr110                    & 16                                & 2946                          & 10                                        & 1159                       \\ \hline 
        lr111                    & 17                                & 2845                          & 10                                        & 1109                       \\ \hline 
        lr112                    & 13                                & 2756                          & 9                                        & 1004                      \\ \hline 
        lr201                    & 9                                & 3295                          & 4                                        & 1253                       \\ \hline 
        lr202                    & 10                                & 3556                          & 3                                        & 1198                       \\ \hline 
        lr203                    & 7                                & 3365                          & 3                                        & 949                       \\ \hline 
        lr204                    & 6                                & 3877                          & 2                                        & 849                       \\ \hline 
        lr205                    & 6                                & 3379                          & 3                                        & 1054                       \\ \hline
        lr206                    & 7                                & 3315                          & 3                                        & 931                       \\ \hline
        lr207                    & 6                                & 3139                          & 2                                        & 903                       \\ \hline
        lr208                    & 5                                & 3322                          & 2                                        & 734                       \\ \hline
        lr209                    & 6                                & 3689                          & 3                                        & 930                       \\ \hline
        lr210                    & 6                                & 3312                          & 3                                        & 964                       \\ \hline
        lr211                    & 5                                & 3241                          & 2                                        & 911                       \\ \hline

    \end{tabular}
    \caption{Results for the \( \CPDPTW \) with 100 tasks (50 customers) on the Li and Lim dataset}
    \label{tab: results cpt}
\end{table}

We managed to solve relatively large instances of \(  \CPDPTW \): instances of 100 tasks of the Li and Lim dataset. We report the results in table \ref{tab: results cpt}
We found that our results cannot compete with the best solutions in literature on this dataset in terms of raw performance (number of vehicles and total distance).
The solutions we produce provide twice the number of vehicle and the distance of the best known solutions. 
However, our approach provides a first end-to-end method to solve the \(  \CPDPTW \) very fast.
We were able to produce all the solutions in less than 40 seconds, and our model can solve several instances in parallel, as well as providing multiple distinct solutions at once.

We witness that our model was not able to grasp the complexity of the time windows, because the model fails to have low number of vehicles solutions
with tight time windows (lr1 instances). 

\begin{figure}
    \centering
    \includegraphics[width=0.7\columnwidth]{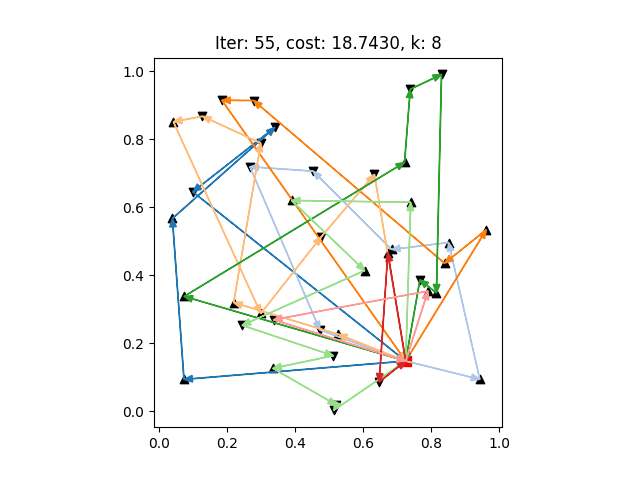}
    \caption{Instance CPTPTW of 20 customers solved by our model, \ding{115}: pick-up, \ding{116}: delivery}
    \label{fig: CPDPTW}
\end{figure}

We trained our model on this setup with 20 customers (40 nodes) and reported an average total distance of 20.94. 

An example of a solution produced by our model coupled to the insertion heuristic is displayed in figure~\ref{fig: CPDPTW}.

\section{Conclusion}
This work shows that it is possible to solve hard planifications problems with end-to-end deep learning models.
While TSP (in other works) and \(  \CVRP \) have been solved very fast and very efficiently for quite some time, it was not possible to extend the resolutions to other problems.
However, these approaches struggle to grasp the complexity of the time-windows constraints. 
Our implementation of the Attention Model for the \( \CVRPTW \) has at least a 20\% optimality gap compared to its classic competitors. 
Meanwhile, other approaches using multi-agent frameworks achieve similar results. 

We show that it is possible to solve the Pick-up and Delivery Problem with Time Windows using an approximate feasibility check using an insertion heuristic.
To our knowledge, we were the first to successfully solve the \(  \CPDPTW \)  with an end-to-end constructivist deep learning model. 

Our approach was successful in creating several solutions fast, which is very valuable in the operational context of SNCF. 
Although performance is not satisfactory yet, the learning approach is promising to produce a lot of different solutions to a hard-constrained problem.
This method seems very promising to tackle dynamic problems, which could be a future research topic.
Using Heterogeneous attention is also a possible extension of this work.

\section*{Acknowledgments}
This work was realized during a research internship at SNCF and fully supported by SNCF, DTIPG, MEV department, MOD group.

\bibliographystyle{unsrt}
\bibliography{deep_RL_CPDPTW}

\end{document}